\def\paperlanguage{}
\pdfoutput=1 % for arxiv

\documentclass[letterpaper, 10 pt, conference]{ieeeconf}  % Comment this line out if you need a4paper

\usepackage{bm}
\usepackage{cite}
\usepackage{flushend}
\include{preamble}

\IEEEoverridecommandlockouts                              % This command is only needed if
\title{\LARGE \textbf
  {
    \switchlanguage%
    {%
      Online Self-body Image Acquisition Considering Changes in Muscle Routes Caused by Softness of Body Tissue for Tendon-driven Musculoskeletal Humanoids
    }%
    {%
      筋骨格ヒューマノイドにおける身体組織の柔軟性による\\筋経路変化を考慮した逐次的自己身体像の獲得
    }%
  }
}

\author{Kento Kawaharazuka, Shogo Makino, Masaya Kawamura,\\ Ayaka Fujii, Yuki Asano, Kei Okada and Masayuki Inaba% <-this % stops a space
  \thanks{Authors are with Department of Mechano-Informatics, Graduate School of Information Science and Technology, The University of Tokyo, 7-3-1 Hongo, Bunkyo-ku, Tokyo, 113-8656, Japan.
    {\texttt\small [kawaharazuka, makino, kawamura, a-fujii, asano, k-okada, inaba]@jsk.t.u-tokyo.ac.jp}
  }
}
\begin{document}

\maketitle
\thispagestyle{empty}
\pagestyle{empty}

%%%%%%%%%%%%%%%%%%%%%%%%%%%%%%%%%%%%%%%%%%%%%%%%%%%%%%%%%%%%%%%%%%%%%%%%%%%%%%%%
\begin{abstract}
  \switchlanguage%
  {%
    Tendon-driven musculoskeletal humanoids have many benefits in terms of the flexible spine, multiple degrees of freedom, and variable stiffness.
    At the same time, because of its body complexity, there are problems in controllability.
    First, due to the large difference between the actual robot and its geometric model, it cannot move as intended and large internal muscle tension may emerge.
    Second, movements which do not appear as changes in muscle lengths may emerge, because of the muscle route changes caused by softness of body tissue.
    To solve these problems, we construct two models: ideal joint-muscle model and muscle-route change model, using a neural network.
    We initialize these models by a man-made geometric model and update them online using the sensor information of the actual robot.
    We validate that the tendon-driven musculoskeletal humanoid Kengoro is able to obtain a correct self-body image through several experiments.
  }%
  {%
    筋骨格ヒューマノイドには柔軟な脊椎構造や多自由度, 可変剛性の実現や球関節の適用という観点から多くの利点が存在する.
    しかし同時に, その複雑な身体故に制御性において大きな問題が生じる
    1つ目は, 実機とモデルの間の誤差が非常に大きいため, 意図した姿勢になることができず, また, 誤差の大きな拮抗関係によって大きな内力が溜まってしまうことがある.
    2つ目は, 身体組織の柔軟性に起因する筋経路変化によって, 筋長センサの変化には現れないような動作が生じてしまう.
    これら二つの問題を解決するために, 理想的な関節-筋空間モデルと筋経路変化の補償モデルという二つのモデルを構築した.
    これらを人間の作った簡易なモデルから初期化し, その後, 実機のセンサデータを用いてオンラインで逐次的に再学習させることで, 自己身体像の獲得を行った.
    自己身体像を用いた関節角度推定, 重量物体保持実験においてその有効性を確認した.
  }%
\end{abstract}

%%%%%%%%%%%%%%%%%%%%%%%%%%%%%%%%%%%%%%%%%%%%%%%%%%%%%%%%%%%%%%%%%%%%%%%%%%%%%%%%
\section{INTRODUCTION} \label{sec:1}
\begin{figure}[t]
  \centering
  \includegraphics[width=1.0\columnwidth]{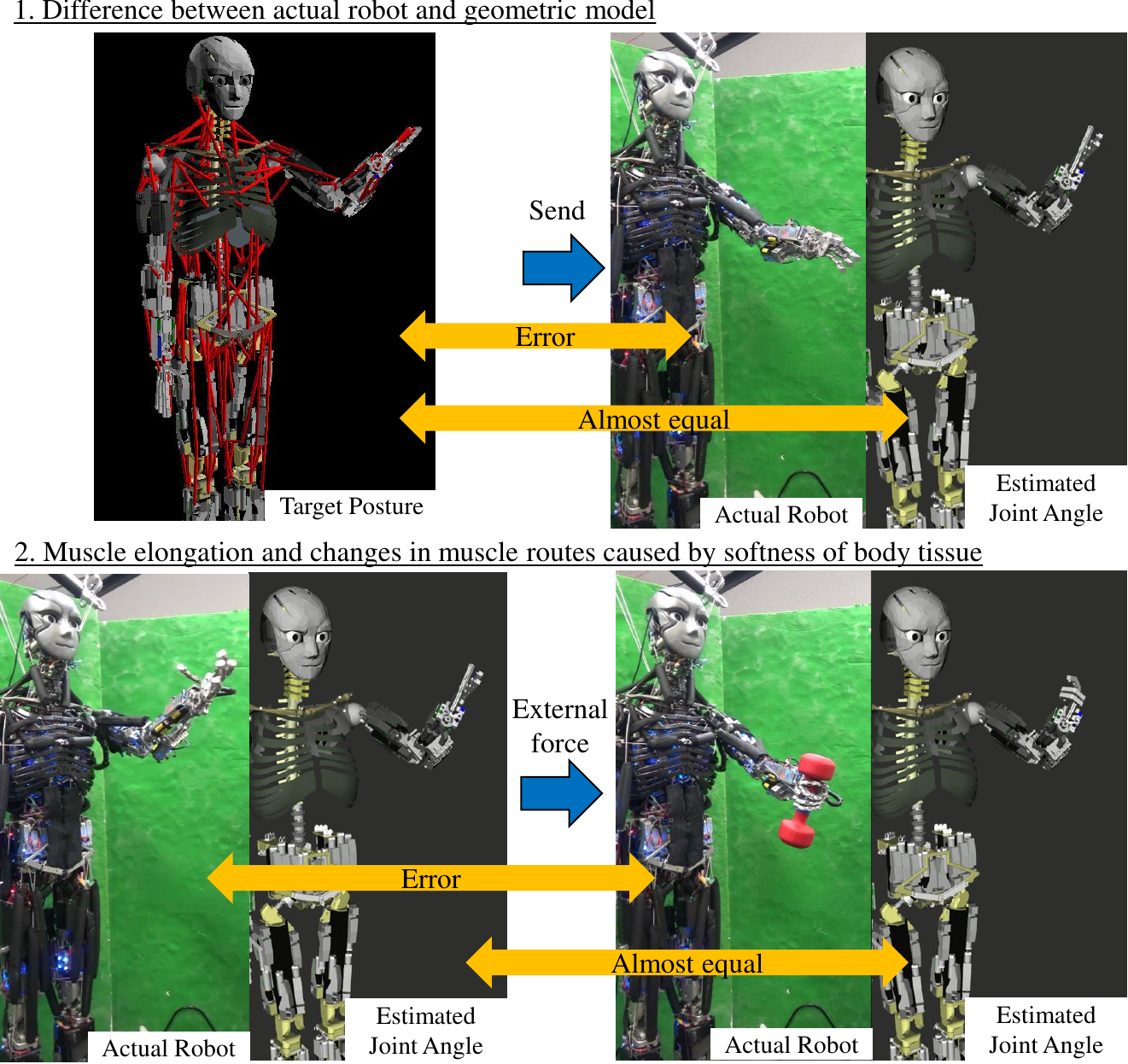}
  \caption{Problems of musculoskeletal humanoids. Upper figure shows that the actual robot cannot move as intended in a simulation environment. Lower figure shows that there are movements which do not appear as changes in muscle lengths.}
  \label{figure:motivation}
\end{figure}

\switchlanguage%
{%
  Tendon-driven musculoskeletal humanoids \cite{ijars2013:nakanishi:approach,artl2013:wittmeier:ecce,humanoids2013:michael:anthrob,humanoids2016:asano:kengoro}, which imitate not only the proportion but also the joint and muscle structures of human beings, have many benefits in terms of the soft spine structure, multiple degrees of freedom (multi-DOF), realization of variable stiffness, and application of ball joints.
  Also, these humanoids are useful for the understanding of the human body system and its applications.

  However, due to complexity of the body structure, there are several challenges in controlling tendon-driven musculoskeletal humanoids.
  The challenges arise from the difficulty to modelize these humanoids compared to the ordinary axis-driven humanoids \cite{icra1998:hirai:asimo}.
  There are mainly two problems (\figref{figure:motivation}).
  First, the model error between the actual robot and its geometric model is very large, and the actual robot cannot move as intended in a simulation environment.
  Usually, we construct its geometric model, check its movements in a simulation environment, and then move the actual robot.
  However, tendon-driven musculoskeletal humanoids have many complex structures, such as multi-DOF joint structure of the spine and redundant muscles covering joints, and it is very difficult to modelize in detail.
  Thus, these humanoids cannot realize the intended joint angles and large internal muscle tension may emerge due to the large model error of antagonistic muscles (the upper figures of \figref{figure:motivation}).
  Second, there are movements which the muscle length sensors cannot measure, due to changes in muscle routes caused by softness of body tissue.
  The muscles are deformed by emerging tension, its routes are changed by external force, and structures such as the rib can deform by muscle tension.
  For example, when tendon-driven musculoskeletal humanoids hold heavy objects, changes in muscle routes which the muscle length sensors cannot measure emerge, and so these humanoids cannot perceive self-body movements (the lower figures of \figref{figure:motivation}).
}%
{%
  人間のプロポーションだけでなく関節構造や筋構造さえも模倣した筋骨格ヒューマノイド\cite{ijars2013:nakanishi:approach,artl2013:wittmeier:ecce,humanoids2013:michael:anthrob,humanoids2016:asano:kengoro}には, 柔軟な脊椎構造や多自由度, 可変剛性の実現や特異点のない球関節の適用という観点から多くの利点がある.
  また, 人間の身体システムの理解, その応用という観点からも有効である.

  しかし同時に, その身体の複雑性ゆえに制御性に関して問題が生じる.
  それらは, 現在普及している軸駆動型ロボットと比較して体の構造を完全にはモデル化できないという点に起因し, 大きく分けて二つの問題がある(図\figref{figure:motivation}).
  一つ目は, 幾何モデルと実機の誤差が大きく, 幾何モデル上で意図した動きが実機では実現できない点である.
  通常ロボットは幾何モデルを作成し, そのモデルをシミュレーション環境で動作させた後に実機を操作するが, 筋骨格ヒューマノイドにおいては, 脊椎などの複雑な多自由度関節構造や, 関節を覆うような経路をとる冗長な筋肉を詳細にモデル化することは難しい.
  よって, 意図した関節角度を実現できない, また, 拮抗筋同士が高張力を発揮し意図しない大きな内力が発生するという問題が生じる.
  二つ目は, 身体組織自体の柔軟性により筋経路が変化することで, 筋長センサの変化には現れない動きが存在するという点である.
  筋は張力を発揮することで変形し, 外力によっても形や経路を変え, また, 肋骨等の構造は変形を伴う.
  よって, 例えば重い物体を持った場合には張力の高まりによって筋長センサの変化として現れない筋経路変化が生じ, 体は大きく動作するもののそれを知覚することはできない(図\figref{figure:motivation}の下段).
}%

\switchlanguage%
{%
  In the previous study \cite{ral2018:kawaharazuka:vision-learning}, we solved the first problem by constructing joint-muscle model (JMM) which expresses the nonlinear relationship between joint angles and muscle lengths using a neural network (NN), and updating it online using the sensor information of actual robot.
  In this study, we expand this method and solve the two problems (\figref{figure:motivation}) simultaneously.
  At first, we initialize the self-body image using a man-made geometric model, and then we update it using the sensor information of the actual robot.
  In the previous study, we expressed self-body image using only ideal joint-muscle model (IJMM).
  However, in this study, we express self-body image using two models: ideal joint-muscle model (IJMM) and muscle-route change model (MRCM), and update them simultaneously.
  By using this method, tendon-driven musculoskeletal humanoids become able to perceive self-body movements which do not appear as changes in muscle lengths, and can move as intended even when external forces are applied.

  This paper is organized as follows.
  In \secref{sec:1}, we stated the motivation and goal of this study.
  In \secref{sec:2}, we will state the overview of this study: the definition of self-body image and the outline of this method.
  In \secref{sec:3}, we will state the initial training method of self-body image using a man-made geometric model.
  In \secref{sec:4}, we will state the online learning method of self-body image using the sensor information of the actual robot.
  In \secref{sec:5}, we will conduct several experiments using this proposed system.
  Finally, in \secref{sec:6}, we will state the conclusion and the future works.
}%
{%
  一つ目の問題に関しては先行研究\cite{ral2018:kawaharazuka:vision-learning}で, 関節角度と筋長の非線形な写像を表す関節-筋空間マッピングをニューラルネットワークで表現し, それを実機の情報を元にオンラインで学習させることで解決を行った.
  本研究では先行研究を発展させ, 先に述べた二つの問題を同時に解決する手法について考える.
  解決方法の大枠は先行研究と同様であり, 図\figref{figure:concept}のように, 人間が作成した幾何モデルを用いて自己身体像を初期学習させ, それを実機から得たセンサデータを元にオンラインで修正していくというものである.
  この中で, 先行研究では自己身体像を理想的な関節-筋空間モデルのみで表現していたのに対して, 本研究では自己身体像を理想的な関節-筋空間モデルと筋経路変化に対する補償モデルという二つのモデルとして構築し, それらを同時に更新していく.
  これによって, 筋経路変化に現れない動作の知覚や, 外力が加わった状態での意図した動作が可能となる.
}%

%\begin{figure}[t]
%  \centering
%  \includegraphics[width=1.0\columnwidth]{figs/concept}
%  \caption{The outline of this research.}
%  \label{figure:concept}
%\end{figure}

%%%%%%%%%%%%%%%%%%%%%%%%%%%%%%%%%%%%%%%%%%%%%%%%%%%%%%%%%%%%%%%%%%%%%%%%%%%%%%%%
\section{Overview of This Study} \label{sec:2}
\switchlanguage%
{%
  First, we will discuss how we define self-body image in this study and what kind of models can express the self-body image.
  Second, we wll state the overview of the proposed system in this study.
}%
{%
  まず, 本研究で定義する自己身体像とは何か, それを表現するためにはどのようなモデルがあり得るかについて議論する.
  その後, システム全体の概要を述べる.
}%

\subsection{Self-Body Image} \label{subsec:self-body-image}
\switchlanguage%
{%
  Although there are several definitions of self-body image, in this study, we define the correct self-body image as ``the state in which the robot can realize intended joint angles".
  In the case of ordinary axis-driven humanoids, they have encoders in each joint and it is possible to realize the intended joint angles by using the feedback control against the error between the value of encoders and intended joint angles.
  However, as stated in \secref{sec:1}, tendon-driven musculoskeletal humanoids have many complex structures and do not have encoders in each joint, so it is difficult to modelize these humanoids in detail and realize the intended joint angles.
  In the previous study \cite{ral2018:kawaharazuka:vision-learning}, we constructed joint-muscle mapping (JMM) which calculates muscle lengths $\bm{l}_{target}$ realizing the intended joint angles $\bm{\theta}_{target}$ using a neural network (NN), and then, updated it online using the sensor information of the actual robot in order to obtain a correct self-body image.
  \begin{align}
    \bm{l}_{target} = \bm{f}(\bm{\theta}_{target})
  \end{align}

  However, this method does not consider the muscle route changes caused by softness of body tissue, and so the robot cannot realize the intended joint angles in such situations as when external force is applied by holding a heavy object.
  Although these muscle-route changes do not appear as changes in muscle lengths, these appear as changes in muscle tensions $\bm{T}$.
  Thus, in this study, we express self-body image not by joint-muscle mapping but by joint-tension-length mapping (JTLM) as shown below.
  \begin{align}
    \bm{l}_{target} = \bm{f}(\bm{\theta}_{target}, \bm{T})
  \end{align}

  Next, we discuss how we should modelize the function $\bm{f}$.
  First, in order to update the self-body image online using the sensor information of the actual robot, in this study, we express the function $\bm{f}$ not by using table-search \cite{icra2010:nakanishi:table} and polynomial regression \cite{humanoids2015:ookubo:muscle-learning} which many previous studies use, but by using a NN.
  In this case, it is the simplest and easiest way to express the function $\bm{f}$ as a NN in which the inputs are $\bm{\theta}$ and $\bm{T}$, and the ouput is $\bm{l}$.
  However, there is a problem.
  When we update the NN online, we backpropagate the error between the output of the function $\bm{f}$ and the sensor value of the actual robot.
  In doing so, we cannot decide if the backpropagated error is one between the actual robot and its geometric model, or one resulting from muscle-route changes caused by softness of body tissue.
  For this reason, even when the error is caused by softness of body tissue, self-body image has often been incorrectly updated in the direction to reduce the error between the actual robot and its geometric model.
  So, in this study, we divide self-body image into two models: ideal joint-muscle model (IJMM, $\bm{f}_{ideal}$) in which no external forces are applied, and muscle-route change model (MRCM, $\bm{g}$) which compensates for the muscle-route changes caused by softness of body-tissue.
  \begin{align}
    \bm{l}_{target} = \bm{f}_{ideal}(\bm{\theta}_{target}) + \bm{g}(\bm{\theta}_{target}, \bm{T}) \label{equation:model}
  \end{align}
  By the construction of this model, we can choose which model to backpropagate the error to, and obtain self-body image stably and correctly.
  In this study, we each express these two models as a NN and propose the system to update them online.
}%
{%
  自己身体像というと様々なレベルが存在するが, 本研究では「意図した関節角度を実現できる」ということを自己身体像が正しい状態として定義する.
  通常の軸駆動型ヒューマノイドであれば現在の関節エンコーダ値と意図したエンコーダ値の差分に対してフィードバックをかければ良いが, 筋骨格ヒューマノイドにおいては第一章で述べたように身体構造を詳細にモデル化できないため難しい.
  先行研究\cite{ral2018:kawaharazuka:vision-learning}では, 筋骨格ヒューマノイドにおいて意図した関節角度$\bm{\theta}$を実現する筋長$\bm{l}_{target}$を計算する関節-筋空間マッピング
  \begin{align}
    \bm{l}_{target} = \bm{f}(\bm{\theta}_{target})
  \end{align}
  をニューラルネットワークで表現し, これを実機データを用いて更新し, 正しい自己身体像を求めた.

  しかしこれには身体組織の柔軟性による筋経路変化が含まれておらず, 重い物体を持つなど, 外力を加えられた状態で意図した関節角度を実現することはできない.
  この筋経路変化は筋長センサの変化としては現れないが, 筋張力$\bm{T}$の変化として現れる.
  そこで, 本研究では関節-筋張力-筋長空間マッピングを自己身体像とする.
  \begin{align}
    \bm{l}_{target} = \bm{f}(\bm{\theta}_{target}, \bm{T})
  \end{align}

  次に, この関数$\bm{f}$をどのようにモデル化すれば良いかについて考える.
  実機のデータからオンラインでモデル修正を容易にするために, 本研究では多くの先行研究で用いられていたテーブルサーチ\cite{icra2010:nakanishi:table}や多項式近似\cite{humanoids2015:ookubo:muscle-learning}ではなく, ニューラルネットワークを用いて関数$\bm{f}$を表現する.
  このとき, 入力を$\bm{\theta}$と$\bm{T}$, 出力を$\bm{l}$としたニューラルネットワークによって関数$\bm{f}$を表現するのが最も単純である.
  しかし, これをオンラインで修正する場合, 現在の関数$\bm{f}$の出力と実機のセンサデータ間の誤差を逆伝播することになるが, その誤差が幾何モデルと実機の間における誤差として, または身体組織の柔軟性による筋経路変化における誤差として, どちらとして逆伝播されるかはわからない.
  そのため, その誤差が身体組織の柔軟性に起因するものであるにもかかわらず, 幾何モデルと実機の間の誤差を解消する方向に学習してしまうことがしばしば起きた.
  そこで本研究では, 自己身体像を理想的な関節-筋空間モデル$\bm{f}_{ideal}$と身体組織の柔軟性による筋経路変化の補償モデル$\bm{g}$という二つに分割して表すこととする.
  \begin{align}
    \bm{l}_{target} = \bm{f}_{ideal}(\bm{\theta}_{target}) + \bm{g}(\bm{\theta}_{target}, \bm{T})
    \label{equation:model}
  \end{align}
  これによって, 誤差を伝播させたいモデルを選ぶことができ, 学習の安定化・より正しい自己身体像の獲得が可能となる.
  この二つのモデルをそれぞれ違ったニューラルネットワークで表現し, それを逐次的に修正していくシステムを本研究では考案する.
}%

\begin{figure}[t]
  \centering
  \includegraphics[width=1.0\columnwidth]{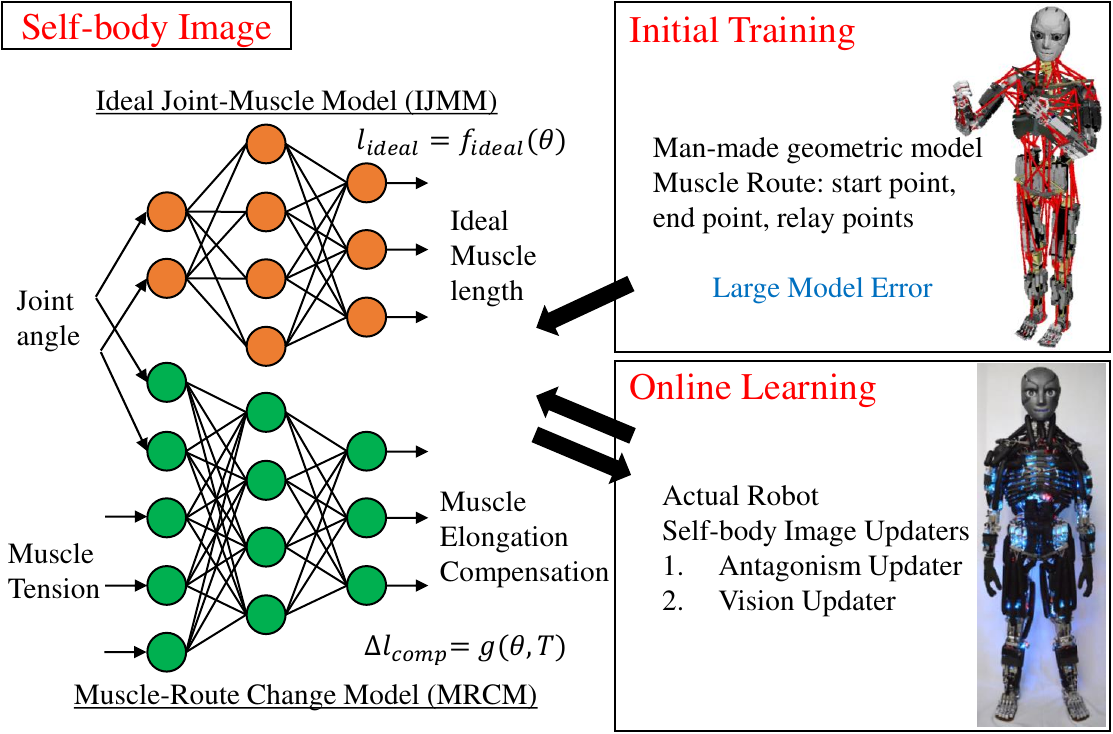}
  \caption{Overview of this system: initial training and online learning.}
  \label{figure:overall-concept}
\end{figure}

\subsection{Overview of This System}
\switchlanguage%
{%
  The overview of the proposed system is shown in \figref{figure:overall-concept}.
  As stated in the previous \secref{subsec:self-body-image}, self-body image is divided into two models: IJMM and MRCM, and each model is expressed using a NN.
  First, as shown in the upper right figure, we construct a data set using a man-made geometric model, train the respective models, and initialize the weight of these NNs.
  Second, as shown in the lower right figure, we update the two models online using the sensor information of the actual robot.
  There are two online updaters: Antagonism Updater and Vision Updater, and the Antagonism Updater updates the antagonism of IJMM, and the Vision Updater updates the IJMM and MRCM based on the information of the vision sensor.

  In terms of high computational cost, these models are divided into body regions such as the neck, scapula, shoulder, and forearm, and to unify these models into one whole body model is a future task.
  As an example, we used a tendon-driven musculoskeletal humanoid Kengoro \cite{humanoids2016:asano:kengoro} in the experiments.
}%
{%
  まず, 図の右上のように人間が作成した幾何モデルを用いて二つのモデルを初期学習し, ある程度のネットワークの重みを構築する.
  ただし, これには実機との間の大きな誤差が含まれている.
  次に, 図の右下のように実機のデータを用いてオンラインでのモデル修正を行うが, ここでは二つの更新則, Antagonism UpdaterとVision Updaterを用い, Antagonism UpdaterはIJMMの拮抗関係の修正を行い, Vision Updaterは視覚から得られた情報を元にIJMMとMRCMを更新していく.
  なお, これらのモデルは計算量の観点から首・肩甲骨・肩・前腕のように部位ごとに別れており, 全身のモデルに拡張するのは今後の課題である.
}%

%%%%%%%%%%%%%%%%%%%%%%%%%%%%%%%%%%%%%%%%%%%%%%%%%%%%%%%%%%%%%%%%%%%%%%%%%%%%%%%%
\section{Initial Training Using a Man-made Geometric Model} \label{sec:3}
\begin{figure}[t]
  \centering
  \includegraphics[width=1.0\columnwidth]{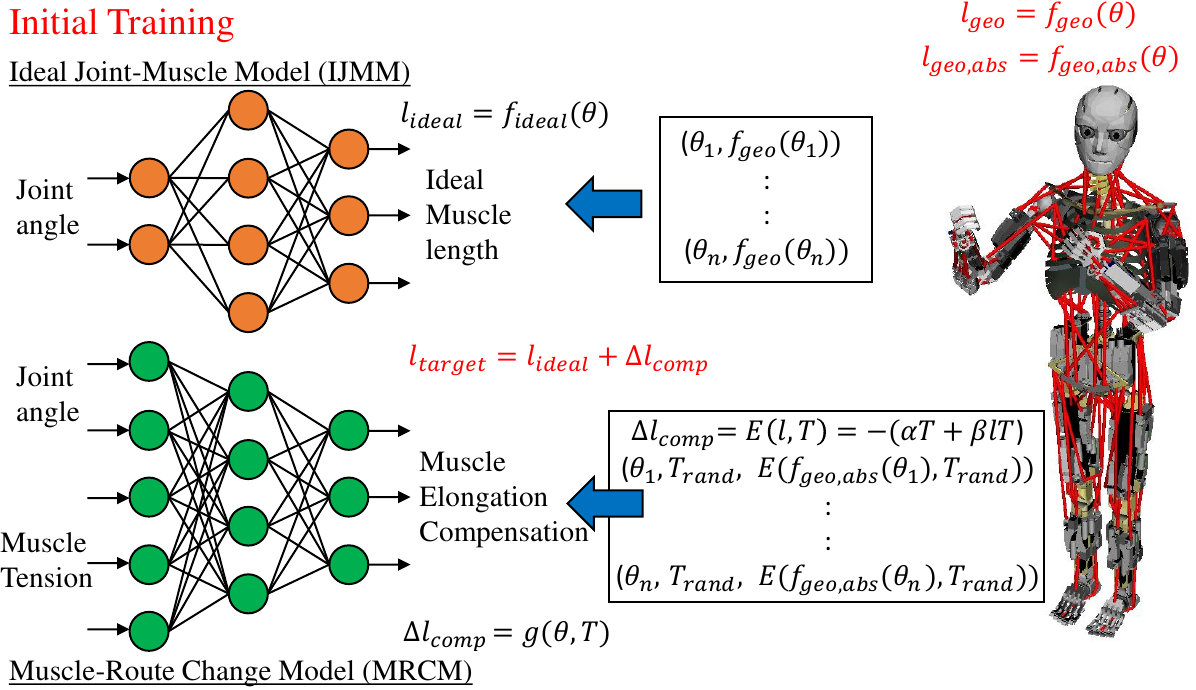}
  \caption{Overview of initial training.}
  \label{figure:initial-training}
\end{figure}

\switchlanguage%
{%
  In this section, we will state the initial training method of IJMM and MRCM (\figref{figure:initial-training}).
}%
{%
  本章では幾何モデルを用いたIJMMとMRCMの初期学習について述べる(図\figref{figure:initial-training}).
}%

\subsection{Ideal Joint-Muscle Model} \label{subsec:ijmm}
\switchlanguage%
{%
  As stated in \equref{equation:model}, IJMM is the model which calculates the target muscle lengths from the target joint angles $\bm{\theta}$, in the ideal case that there are no muscle route changes caused by softness of body tissue.
  First, we make a geometric model of muscle routes in the tendon-driven musculoskeletal humanoid, for example, by expressing the start point, relay point, and end point of the muscle route.
  Also, in this study, we assume that the correct geometric model of the joint structures can be obtained by computer-aided design (CAD).
  Second, we move the joints of the geometric model to various angles, obtain the relative changes in muscle lengths from the initial posture (all joint angles are 0) by using the muscle routes of the geometric model, $f_{geo}$, and construct a data set of joint angles and muscle lengths (the upper right figure of \figref{figure:initial-training}).
  Finally, we train IJMM by using this data set.
}%
{%
  理想関節-筋空間モデル(IJMM)は式(\ref{equation:model})にあるように, 身体組織の柔軟性に起因する筋経路に変化のない理想的な状態における, ターゲットとなる関節角度$\bm{\theta}$からターゲットとなる筋長を求めるモデルである.
  まず, 筋骨格ヒューマノイドの筋経路に関する幾何モデルを作成するが, 本研究では筋肉一つ一つをその始点・中継点・終止点を元にモデル化したものを作成した.
  また, 本研究では関節構造に関してはCADから正しい幾何モデルが得られていると仮定する.
  次に, この幾何モデルの関節を様々に動かし, 初期状態(関節角度が全て0)からの相対的な筋長変化を得ることで, 関節角度と相対筋長のペアデータを取得する(図\figref{figure:initial-training}の右上).
  その後, そのデータを用いてIJMMを学習させる.
}%

\subsection{Muscle-Route Change Model} \label{subsec:mrcm}
\begin{figure}[t]
  \centering
  \includegraphics[width=1.0\columnwidth]{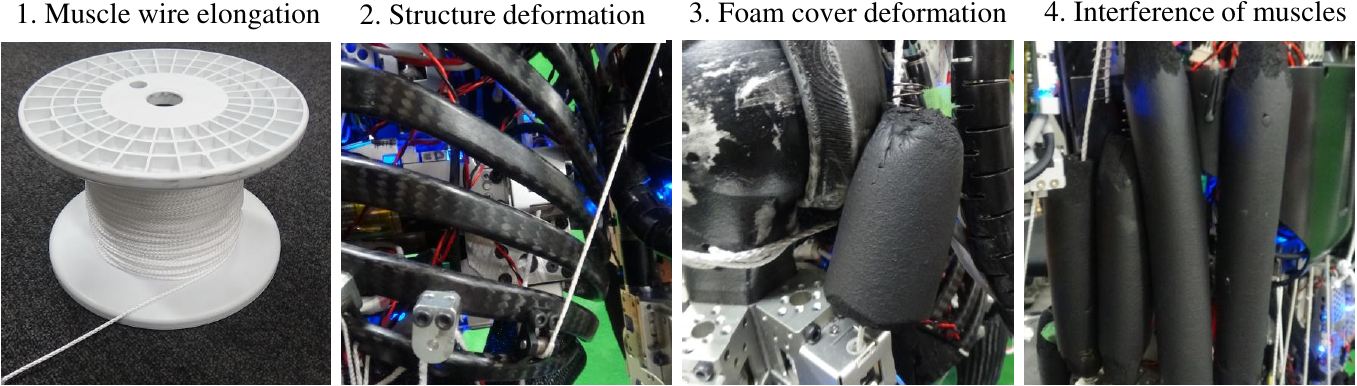}
  \caption{Causes of muscle-route change: muscle wire elongation, structure deformation, foam cover deformation, and interference of muscles.}
  \label{figure:muscle-route-change}
\end{figure}

\switchlanguage%
{%
  As stated in \equref{equation:model}, MRCM is the model which compensates for muscle route changes caused by softness of body tissue, and which calculates the compensation value of muscle length from the target joint angles $\bm{\theta}$ and muscle tensions $\bm{T}$.
  In order to initialize MRCM from a man-made geometric model, we must modelize the cause of muscle route changes to some extent.
  Then, we classify the muscle route changes caused by softness of body tissue into 4 groups (\figref{figure:muscle-route-change}).

  First, there is the elongation of Dyneema ${\Delta}l$ used as muscle wires \cite{iros2015:asano:module}.
  As shown below, the elongation depends on $T$ and $l_{abs}$.
  $l_{abs}$ is calculated from $\bm{\theta}$, so ${\Delta}l$ depends on $T$ and $\bm{\theta}$.
  \begin{align}
    T &= k{\Delta}l/l_{abs}
  \end{align}
  where $k$ is the spring constant of Dyneema per unit length, $l_{abs}$ is the absolute length of the muscle.
  Also, we refer to variables which do not have the subscript $abs$ as the relative muscle lengths from initial posture.
  Second, there is the deformation of structures such as the rib, and this basically depends on muscle tension $T$.
  Third, because the muscles of the tendon-driven musculoskeletal humanoid Kengoro \cite{humanoids2016:asano:kengoro} have soft foam covers around the muscle wires in order to achieve softness of contact, and the muscle route changes occur when the foam covers deform due to large muscle tension.
  This depends on the size of the foam covers and how the muscles wind around the bone structure, and therefore, this depends on the muscle tension $T$ and the posture $\bm{\theta}$ of the robot.
  Fourth, there is the interference among muscles which have soft foam covers, and this depends on the muscle tensions $\bm{T}$ and the posture $\bm{\theta}$ of the robot.

  In short, MRCM depends on muscle tensions $\bm{T}$ and the posture $\bm{\theta}$ of the robot.
  In this study, in order to initialize MRCM, we use the elongation of muscle wires and deformation of the structure, which are easy to modelize, to calculate muscle lengths which should be compensated, ${\Delta}\bm{l}_{comp}$, as shown below.
  \begin{align}
    {\Delta}\bm{l}_{comp} = -({\alpha}\bm{T} + {\beta}\bm{l}\bm{T}) = -({\alpha}\bm{T} + {\beta}\bm{f}_{geo, abs}(\bm{\theta})\bm{T})
  \end{align}
  where the $\alpha$ is a coefficient for the elongation of muscle wires ($10.0$ in this study), $\beta$ is a coefficient for the deformation of the structure ($0.05$ in this study), and $\bm{f}_{geo, abs}$ is the function which calculates the absolute muscle lengths from the geometric model.
  ${\Delta}\bm{l}_{comp}$ is the muscle lengths which should be compensated, so it is a negative value.
  We change $\bm{\theta}$ and $\bm{T}$ randomly, calculate ${\Delta}\bm{l}_{comp}$, and obtain the data set of joint angles, muscle tensions, and compensation value of muscle lengths (the lower right of \figref{figure:initial-training}).
  Finally, we train MRCM using the data set.
}%
{%
  筋経路変化補償モデル(MRCM)は式(\ref{equation:model})にあるように, 身体組織の柔軟性に起因する筋経路変化を補償するためのモデルであり, ターゲットとなる関節角度$\bm{\theta}$とそのときの筋張力$\bm{T}$から補償分の筋長を求める.
  MRCMを幾何モデルから初期化するためには, 筋経路変化の原因をある程度モデル化しなければならない.
  そこでまず, 身体組織の柔軟性に起因する筋経路変化を4つに分類する(図\figref{figure:muscle-route-change}).

  一つ目は, 筋ワイヤとして用いているDyneemaの伸び${\Delta}l$である.
  これは
  \begin{align}
    T &= k{\Delta}l/l_{abs}
  \end{align}
  であり$T$と$l_{abs}$に依存し, $l_{abs}$は$\bm{\theta}$から求まるため, 最終的に${\Delta}l$は$T$と$\bm{\theta}$に依存する.
  ここで$k$はDyneemaの単位長さあたりのバネ係数, $l_{abs}$はその筋の絶対筋長であり, 以降$abs$と添字がない場合は初期姿勢からの相対筋長とする.
  二つ目は, 骨格の撓みであり, これは基本的には筋張力$T$に比例する.
  三つ目は本研究で用いる筋骨格ヒューマノイド腱悟郎\cite{humanoids2016:asano:kengoro}の筋構造に依存するものだが, 筋ワイヤの周りに柔軟な接触を担保するために発泡性カバーが付いており, これが筋張力の高まりによって潰れることで起こる.
  これは発泡性カバーの大きさや骨格への巻きつき方等が関係し, その筋の筋張力$T$と姿勢$\bm{\theta}$に依存する.
  四つ目はそれら発泡性カバーのついた筋同士の干渉であり, これは干渉し合う筋の筋張力$\bm{T}$と姿勢$\bm{\theta}$に依存する.

  まとめると, MRCMは$\bm{T}$と$\bm{\theta}$に依存することがわかる.
  本研究では, MRCMを初期化するためにモデル化の簡単なDyneemaの伸びと骨格の撓みを用い,
  \begin{align}
    {\Delta}\bm{l}_{comp} = -({\alpha}\bm{T} + {\beta}\bm{l}\bm{T}) = -({\alpha}\bm{T} + {\beta}\bm{f}_{geo, abs}(\bm{\theta})\bm{T})
  \end{align}
  として筋経路変化補償分の筋長${\Delta}\bm{l}_{comp}$を求める.
  ここで, $\alpha$と$\beta$は係数(本研究では10, 0.05を用いる), $\bm{f}_{geo, abs}$は幾何モデルから絶対筋長を求める関数であり, ${\Delta}\bm{l}_{comp}$は伸びを補償する項のため, 全体にマイナスがかかっている.
  そして, $\bm{\theta}$と$\bm{T}$の値を様々に変え, ${\Delta}\bm{l}_{comp}$を求めることで関節角度・筋張力・筋経路変化補償のデータセットを取得する(図\figref{figure:initial-training}の右下).
  その後, そのデータを用いてMRCMを学習させる.
}%

%%%%%%%%%%%%%%%%%%%%%%%%%%%%%%%%%%%%%%%%%%%%%%%%%%%%%%%%%%%%%%%%%%%%%%%%%%%%%%%%
\section{Online Self-body Image Acquisition Using The Actual Robot Data} \label{sec:4}
\begin{figure}[t]
  \centering
  \includegraphics[width=1.0\columnwidth]{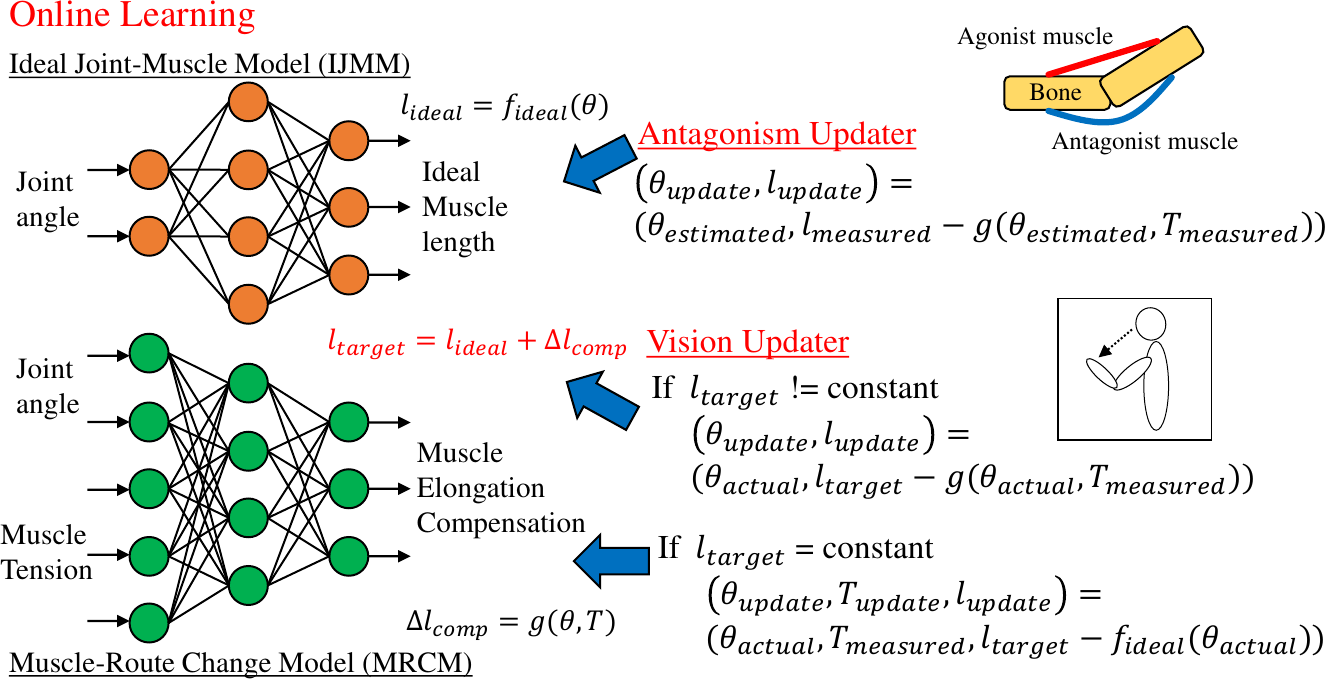}
  \caption{Overview of online learning: Antagonism Updater and Vision Updater.}
  \label{figure:online-learning}
\end{figure}

\switchlanguage%
{%
  In this section, we will explain the components of online learning: Antagonism Updater and Vision Updater, which are necessary to update IJMM and MRCM online from the sensor information of the actual robot (\figref{figure:online-learning}).
}%
{%
  本章では, 実機のセンサデータからIJMMとMRCMをオンラインで修正していくためのコンポーネント, Antagonism Updater, Vision Updaterについてそれぞれ述べる(図\figref{figure:online-learning}).
}%

\subsection{Online Learning}
\switchlanguage%
{%
  We will explain the estimation method of joint angles using muscle length, the estimation method of the actual joint angles using the vision sensor, and the learning methods of IJMM and MRCM, which are necessary to update IJMM and MRCM online.

  First, we will state the method of joint angle estimation.
  Tendon-driven musculoskeletal humanoids have ball joints and flexible spine structures, so encoders cannot be installed and joint angles need to be estimated from the changes in muscle lengths.
  We use the method of Ookubo, et al. \cite{humanoids2015:ookubo:muscle-learning} using Extended Kalman Filter (EKF) for the joint angle estimation.
  In the method, we need JMM, $\bm{f}_{geo}(\bm{\theta})$, of the geometric model and the muscle jacobian, $G_{geo}(\bm{\theta})$, which is the differentiated value of JMM.
  In this study, we substitute the self-body image, $\bm{f}_{ideal}(\bm{\theta})+\bm{g}(\bm{\theta}, \bm{T})$, and its differentiated value for the above JMM and muscle jacobian.
  Also, although Ookubo, et al. used the sensor value of muscle length, $\bm{l}_{measured}$, as the observed value, we use the muscle length we send to the actual robot, $\bm{l}_{target}$, instead.

  Second, we will state the joint angle estimation method of the actual robot, $\bm{\theta}_{actual}$, from the vision sensor.
  The robot can obtain the relative position and posture of the hand from the eye, $P_{vision}$, by looking at the AR marker attached to the hand from the RGB camera attached to the head.
  Then, we obtain the joint angles of the actual robot $\bm{\theta}_{actual}$ by solving the Inverse Kinematics (IK) of the robot, where the initial joint angles are the estimated joint angles stated above, $\bm{\theta}_{est}$, the target position and posture is $P_{vision}$, and the links are the neck, thorax, collarbone, scapula, humerus, and forearm.
  \begin{align}
    \bm{\theta}_{actual} = IK(\bm{\theta}_{initial}=\bm{\theta}_{est}, \bm{P}_{target}=\bm{P}_{vision})
  \end{align}
  We have shown the effectiveness of online learning using these joint angles $\bm{\theta}_{actual}$ in a previous study \cite{ral2018:kawaharazuka:vision-learning}.

  Third, we will state the learning method of IJMM.
  When we update IJMM, we need a data set of joint angles and muscle lengths $(\bm{\theta}_{update}, \bm{l}_{update})$.
  However, if we use only this data set, over-fitting to the given the data set can occur.
  Taking into consideration that the joint angles and muscle lengths are 0 at the initial pose, and that the areas of the model except around the obtained data set should not be updated, and then, we update IJMM using the data set shown below $D$ as a minibatch.
  \begin{align}
    \bm{D} = \{(\bm{\theta}_{update},\bm{l}_{update}), (\bm{0}, \bm{0}), {(\bm{\theta}_{rand}, \bm{f}_{ideal}(\bm{\theta}_{rand}))}_{1\textrm{...}N}\}
  \end{align}
  where $\bm{\theta}_{rand}$ is randomized joint angles, and $N$ is the number of the $\bm{\theta}_{rand}$.

  Fourth, we will state the learning method of MRCM.
  When we update MRCM, we need a data set of joint angles, muscle tensions, and compensation values of muscle lengths, $(\bm{\theta}_{update}, \bm{T}_{update}, \Delta\bm{l}_{update})$.
  The concept of the update is the same as in IJMM.
  Taking into consideration that the compensation values of muscle lengths are 0 when the muscle tensions are 0, and that the change of the compensation value is small when the posture of the robot changes, we update MRCM using the data set $D$ shown below as a minibatch.
  \begin{align}
    \begin{split}
      \bm{D} = \{(\bm{\theta}_{update}, \bm{T}_{update}, \Delta\bm{l}_{update}), {(\bm{\theta}_{rand}, \bm{0}, \bm{0})}_{1\textrm{...}M}, \\{(\bm{\theta}_{around}, \bm{T}_{update}, \Delta\bm{l}_{update})}_{1\textrm{...}N}\}
    \end{split}
  \end{align}
  where $\bm{\theta}_{around}$ is the randomized joint angles around $\bm{\theta}_{update}$.
}%
{%
  オンラインでIJMM, MRCMを修正していくために必要な, 関節角度推定値, 視覚からの実機関節角度の推定, IJMMの学習方法, MRCMの学習方法について述べる.

  まず関節角度推定値であるが, 筋骨格ヒューマノイドは人体を模した球関節や脊椎が多いため関節にエンコーダを入れることができず, 筋長の変化から関節角度を推定する必要がある.
  そのために本研究では大久保らのEKF(Extended Kalman Filter)を使った方法を用いる\cite{humanoids2015:ookubo:muscle-learning}.
  大久保らの手法では幾何モデル上での関節-筋空間マッピング$\bm{f}_{geo}(\bm{\theta})$とその微分である筋長ヤコビアン$G_{geo}(\bm{\theta})$を用いていたが, 本研究ではそれを自己身体像$\bm{f}_{ideal}(\bm{\theta})+\bm{g}(\bm{\theta}, \bm{T})$とその微分に置き換えて用いている.
  また, 大久保らの手法では筋長センサの値$\bm{l}_{measured}$を入力としていたのに対し, 本研究では実機に送っている筋長$\bm{l}_{target}$を用いている.

  次に視覚からの実機関節角度$\bm{\theta}_{actual}$の推定についてである.
  本研究ではロボットの視覚であるRGBカメラから手先についたARマーカを見ることで目から手先への相対位置姿勢$P_{vision}$得る.
  初期姿勢を先ほど述べた関節角度推定値$\bm{\theta}_{est}$, ターゲットとなる相対位置姿勢を$P_{vision}$として目から首・胸郭・鎖骨・肩甲骨・上腕・前腕をリンクとした逆運動学(IK)を解くことで現在の実機の関節角度推定を行う.
  \begin{align}
    \bm{\theta}_{actual} = IK(\bm{\theta}_{initial}=\bm{\theta}_{est}, \bm{P}_{target}=\bm{P}_{vision})
  \end{align}
  この$\bm{\theta}_{actual}$を用いることによる実機学習の有効性は, 先行研究\cite{ral2018:kawaharazuka:vision-learning}において示されている.

  次に, IJMMの学習方法を述べる.
  IJMMを更新する際は関節角度と筋長のデータ$(\bm{\theta}_{update}, \bm{l}_{update})$が必要となる.
  しかし, このデータのみでIJMMを更新すると得られたデータの部分のみ過学習してしまうことが想定されるため, 初期姿勢では関節角度は0, 筋長も0であるということ, また, 得られたデータ周辺以外の部分は元のモデルから変えるべきではないことを考慮し, 以下のデータ$D$を一度にミニバッチとして学習させる.
  \begin{align}
    \bm{D} = \{(\bm{\theta}_{update},\bm{l}_{update}), (\bm{0}, \bm{0}), {(\bm{\theta}_{rand}, \bm{f}_{ideal}(\bm{\theta}_{rand}))}_{1\textrm{..}N}\}
  \end{align}
  ここで, $\bm{\theta}_{rand}$はランダムな関節角度であり, $N$はランダムなデータの数である.

  最後に, MRCMの学習方法を述べる.
  MRCMを更新する際は関節角度と筋張力と補正分の筋長のデータ$(\bm{\theta}_{update}, \bm{T}_{update}, \Delta\bm{l}_{update})$が必要となる.
  これは先程と同様, 筋張力が0の場合は筋経路変化補償も0であるということ, 姿勢変化による補正分の筋長変化は少ないということを用い, 以下のデータ$D$を一度にミニバッチとして学習させる.
  \begin{align}
    \begin{split}
      \bm{D} = \{(\bm{\theta}_{update}, \bm{T}_{update}, \Delta\bm{l}_{update}), {(\bm{\theta}_{rand}, \bm{0}, \bm{0})}_{1\textrm{..}M}, \\{(\bm{\theta}_{around}, \bm{T}_{update}, \Delta\bm{l}_{update})}_{1\textrm{..}N}\}
    \end{split}
  \end{align}
  ここで, $\bm{\theta}_{around}$は$\bm{\theta}_{update}$周辺のランダムな関節角度である.
}%

\subsection{Antagonism Updater}
\switchlanguage%
{%
  As stated in \secref{sec:1}, due to the model error between the actual robot and its geometric model, not only is the robot unable to realize the intended joint angles, but also the large internal muscle tension can emerge because of the wrong antagonistic relationship.
  The method to modify the antagonism is the Antagonism Updater, and this updates only IJMM.
  The rule of the update is shown as below.
  \begin{align}
    (\bm{\theta}_{update}, \bm{l}_{update}) = (\bm{\theta}_{est}, \bm{l}_{m}-\bm{g}(\bm{\theta}_{est}, \bm{T}_{m}))
  \end{align}
  where $\bm{l}_{m}$ is the muscle lengths obtained from the encoders attached to the muscle actuators, $\bm{T}_{m}$ is the current muscle tensions.
  The subscript $m$ stands for $measured$.
  Also, the update is executed only at the static state (when the change of $\bm{\theta}_{est}$ is small) and only when $\bm{\theta}_{update}$ has changed to some extent from the $\bm{\theta}_{update}$ updated previously.

  We will state the reason why the antagonism becomes better by this Antagonism Updater.
  We use the muscle stiffness control \cite{robio2011:shirai:control} when we send target muscle length to the actual robot as shown below.
  \begin{align}
    \bm{T}_{target} = \bm{T}_{bias} + \textrm{max}\{0, K_{stiff}(\bm{l}-\bm{l}_{target})\}
  \end{align}
  In this control, the smaller the muscle stiffness $K_{stiff}$ is, the more the $\bm{l}$ does not follow $\bm{l}_{target}$, permitting the error.
  So if we move the robot by the unfeasible antagonism, large internal muscle tension emerges, but the muscles settle down to a certain state by permitting the error.
  By updating IJMM using the current estimated joint angles and the feasible muscle lengths, the antagonism becomes correct and large internal muscle tension does not emerge.
  Also, although we stated that the muscle length of the actual robot does not follow the target muscle length because of muscle stiffness control, the compensation value of the error is considered into MRCM in addition to the 4 groups of \figref{figure:muscle-route-change}.
}%
{%
  第一章で述べたように, 幾何モデルと実機の間の誤差によって意図した姿勢にならないというだけでなく, 間違った拮抗関係によって大きな内力が高まってしまう.
  この拮抗関係を修正するのがAntagonism Updaterであり, IJMMのみを修正する.
  更新則は
  \begin{align}
    (\bm{\theta}_{update}, \bm{l}_{update}) = (\bm{\theta}_{est}, \bm{l}_{m}-\bm{g}(\bm{\theta}_{est}, \bm{T}_{m}))
  \end{align}
  である.
  ここで, $\bm{l}_{m}$は筋アクチュエータについたエンコーダから得た筋長, $\bm{T}_{m}$は現在の筋張力であり, $m$は$measured$を表す.
  また, 更新は静止状態($\bm{\theta}_{est}$の動きが小さい)で, 前回更新した$\bm{\theta}_{update}$から動作があった場合に行う.

  この更新則によって拮抗関係が働く理由を述べる.
  筋長を送る際には筋剛性制御\cite{robio2011:shirai:control}
  \begin{align}
    \bm{T}_{target} = \bm{T}_{bias} + \textrm{max}\{0, K_{stiff}(\bm{l}-\bm{l}_{target})\}
  \end{align}
  を用いており, 筋剛性$K_{stiff}$の値が小さいほど$\bm{l}$が$\bm{l}_{target}$に追従せず誤差を許容するため, 実現可能でない拮抗関係で動作をさせた場合に大きな内力が溜まるものの, 誤差を許容し実現可能な状態に落ち着く.
  そのときの関節角度推定値と実現可能となった筋長を用いてIJMMを更新することで, 拮抗関係が正しく, 内力が溜まり過ぎない拮抗関係になっていく.
  また, 筋剛性制御によって実際の筋長がターゲットとなる筋長に追従しないと述べたが, これに対する補償項は図\figref{figure:muscle-route-change}の4つに加えてMRCMに考慮されることになる.
}%

\subsection{Vision Updater}
\switchlanguage%
{%
  Vision Updater updates both IJMM and MRCM using the estimated joint angles of the actual robot, $\bm{\theta}_{actual}$, obtained from vision sensor.
  The update rule of IJMM is as shown below.
  \begin{align}
    If \; \; \bm{l}_{target} !=& constant&  \\
    (\bm{\theta}_{u}, \bm{l}_{u}) =& (\bm{\theta}_{actual}, \bm{l}_{target}-\bm{g}(\bm{\theta}_{actual}, \bm{T}_{m}))
  \end{align}
  The update rule of MRCM is as shown below.
  \begin{align}
    If \; \; \bm{l}_{target} &= constant& \\
    (\bm{\theta}_{u}, \bm{T}_{u}, \bm{l}_{u}) &= (\bm{\theta}_{actual}, \bm{T}_{m}, \bm{l}_{target}-\bm{f}_{i}(\bm{\theta}_{actual}))
  \end{align}
  where we abbreviated $update$ and $ideal$ as $u$ and $i$.
  Also, the update is executed only at the static state (when the change of $\bm{\theta}_{est}$ is small) and only when $\bm{\theta}_{update}$ has changed to some extent from the $\bm{\theta}_{update}$ updated previously.

  In the update rule, $\bm{l}_{target} = constant$ means that the target muscle lengths do not change, so we do not send operating commands to the robot.
  When we do not send operating commands and $\bm{\theta}_{update}$ changes from the previous value, external force is applied, and so we update MRCM, which expresses the compensation value of the muscle length by the applied force.
  To the contrary, when $\bm{l}_{target} != constant$, we update IJMM.
  Thus, we fix one model: IJMM or MRCM, and update the other model.
  However, in the case that we update IJMM when external force is applied and operating commands are sent, the error which should be reflected to MRCM is reflected to IJMM.
  When we update IJMM, we need to check that external force is not applied to the contact sensors.

  Practically, it is better that we update IJMM by Antagonism Updater and Vision Updater first, and after IJMM becomes correct, we update MRCM by applying external forces.
}%
{%
  Vision Updaterでは, 視覚から得た実機の関節角度$\bm{\theta}_{actual}$を用いて, IJMM, MRCMの両方を更新していく.
  IJMMの更新則は
  \begin{align}
    If \; \; \bm{l}_{target} != constant&  \\
    (\bm{\theta}_{u}, \bm{l}_{u}) = (\bm{\theta}_{actual},& \bm{l}_{target}-\bm{g}(\bm{\theta}_{actual}, \bm{T}_{m}))
  \end{align}
  MRCMの更新則は
  \begin{align}
    If \; \; \bm{l}_{target} = constant& \\
    (\bm{\theta}_{u}, \bm{T}_{u}, \bm{l}_{u}) = (\bm{\theta}_{actual},& \bm{T}_{m}, \bm{l}_{target}-\bm{f}_{i}(\bm{\theta}_{actual}))
  \end{align}
  である(添字$update$を$u$, $ideal$を$i$と省略).
  また, 更新は静止状態($\bm{\theta}_{est}$の動きが小さい)で, 前回更新した$\bm{\theta}_{update}$から動作があった場合に行う.

  $\bm{l}_{target} = constant$とは実機に送る筋長に変化がない, つまり, 動作指令を行っていないことを指す.
  動作指令を行っていないときに前回更新した$\bm{\theta}_{update}$から変位があった場合には外力が働いているということであり, このときに外力による筋経路変化補償であるMRCMを更新する.
  逆に, $\bm{l}_{target} != constant$のときにはIJMMを更新する.
  つまり, 片方のモデルを固定してもう片方のモデルを正しくしていくのである.
  ただし, 外力が加わった状態で動作指令を送りIJMMを学習した場合, MRCMに反映されるべき筋長がIJMMに反映されてしまうため, 他に, 手の力覚に力が加わっていない等の条件が必要である.
  また実運用上, まずはAntagonism UpdaterとVision UpdaterによってIJMMを更新し, IJMMが正しくなった後に外力を加えてMRCMを正しくすることが望ましい.
}

\begin{figure}[htb]
  \centering
  \includegraphics[width=1.0\columnwidth]{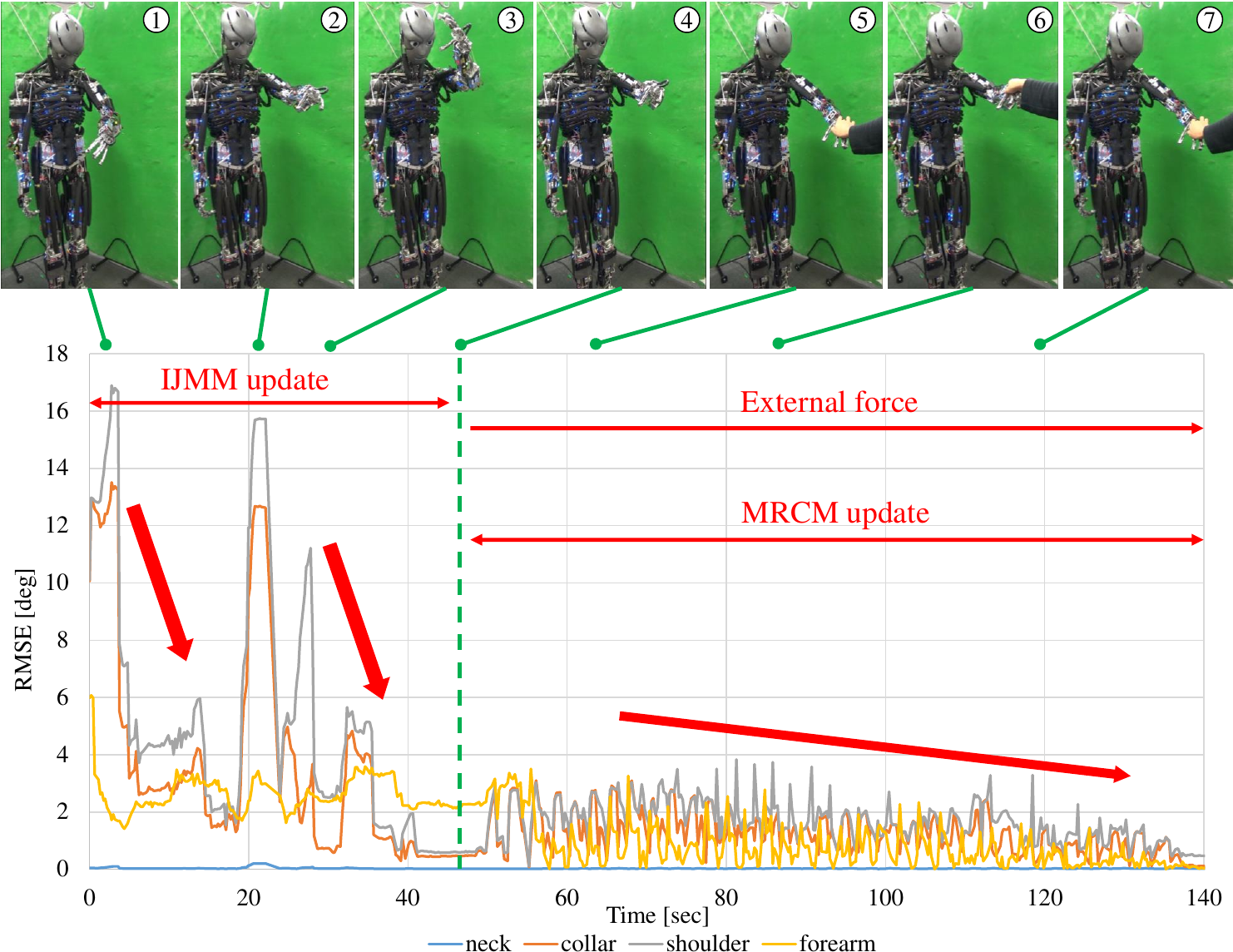}
  \caption{Experiment of online learning: upper figure shows the appearance of this experiment and lower graph shows the Root Mean Squared Error (RMSE) of the difference between the actual and estimated joint angles ($\bm{\theta}_{actual}-\bm{\theta}_{est}$).}
  \label{figure:learning-experiment}
\end{figure}

%%%%%%%%%%%%%%%%%%%%%%%%%%%%%%%%%%%%%%%%%%%%%%%%%%%%%%%%%%%%%%%%%%%%%%%%%%%%%%%%
\section{Experiments} \label{sec:5}
\switchlanguage%
{%
  In this section, at first, we show the result of online learning and verify its effectiveness.
  Next, we show the effectiveness of joint angle estimation using the learned IJMM and MRCM.
  Finally, we conduct a grasping experiment of a heavy object and verify the effectiveness of the learned IJMM and MRC.
}%
{%
}%

\begin{figure}[htb]
  \centering
  \includegraphics[width=0.90\columnwidth]{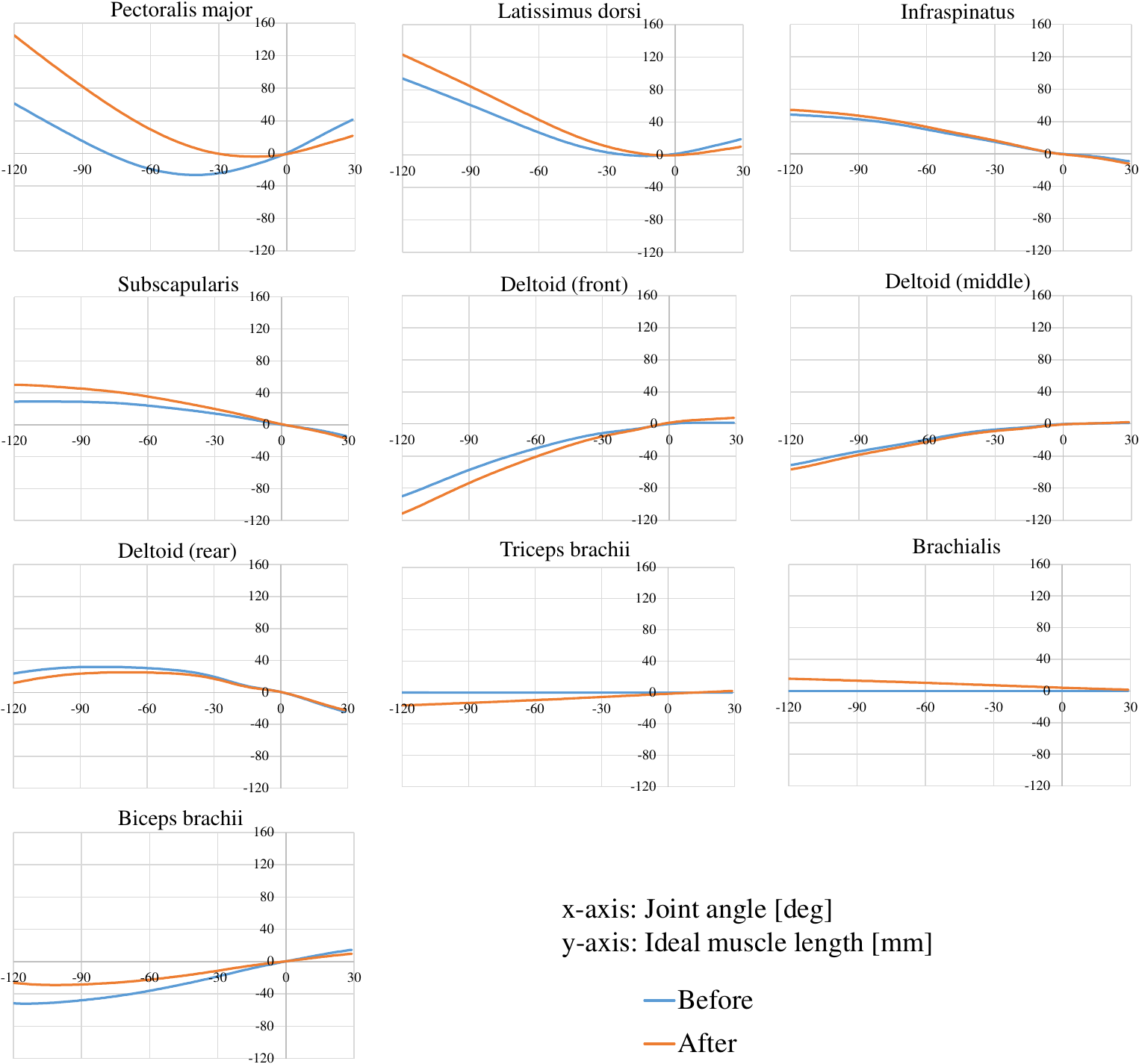}
  \caption{The comparison of IJMM between before and after online learning. We moved the shoulder pitch and measured ideal muscle lengths.}
  \label{figure:learning-model}
\end{figure}

\begin{figure}[htb]
  \centering
  \includegraphics[width=0.90\columnwidth]{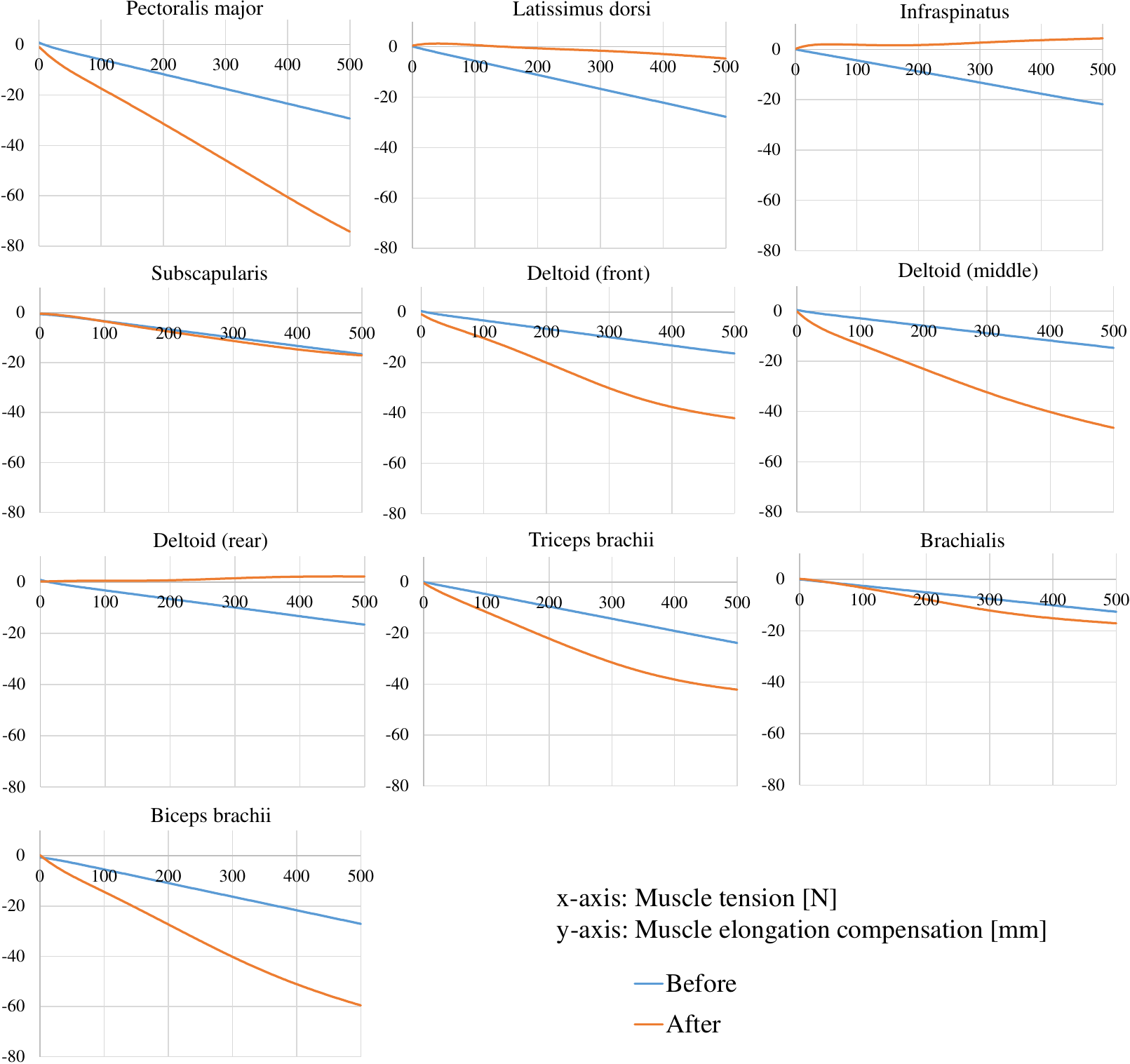}
  \caption{The comparison of MRCM between before and after online learning. At the posture which the joint angles of the shoulder and elbow, (GH-r, GH-p, GH-y, E-p), are (60, -60, -30, -90) [deg] (GH means glenohumeral joint, E means elbow joint, and rpy means roll, pitch, and yaw), we increased muscle tension from 0 N to 500 N and measured the compensation value $\Delta{l}_{comp}$.}
  \label{figure:learning-emodel}
\end{figure}

\subsection{Online Learning}
\switchlanguage%
{%
  We show the result of online learning of IJMM and MRCM.
  The effectiveness of the Antagonism Updater is verified in a previous study \cite{ral2018:kawaharazuka:vision-learning}, so we show the effectiveness of the Vision Updater regarding IJMM and MRCM.
  Also in this study, each NN of IJMM and MRCM has 3 layers: an input layer, a hidden layer (1000 unit), and an output layer of muscle lengths, and the activation function is sigmoid.
  The unit of input joint angles is [rad], the unit of input muscle tension is $T\textrm{[N]}/500$ (about 0$\sim$1), and the unit of output muscle lengths is [mm].

  We show the result of online learning in \figref{figure:learning-experiment}.
  First, we send various joint angles to Kengoro, Kengoro looks at its hand, and we update self-body image using the Vision Updater.
  In this situation, $\bm{l}_{target}$ is not constant and IJMM is updated.
  Next, in the situation that we do not send operating commands, by applying various external forces, we update MRCM.
  In all the movements of this experiment, the Antagonism Updater is also executed.

  We show the transition graph of Root Mean Squared Error (RMSE) of the difference between the actual joint angles obtained from the vision sensor, $\bm{\theta}_{actual}$, and estimated joint angles calculated from the current self-body image, $\bm{\theta}_{est}$, regarding joint groups such as the neck, scapula, shoulder, and forearm in the lower figure of \figref{figure:learning-experiment}.
  During the update of IJMM, RMSE was about 16 deg at first, and at the end, RMSE decreased to about 0$\sim$2 deg.
  Also, RMSE increased once at 20 sec, and this means that the posture moved largely from the posture updated at 0$\sim$5 sec and self-body image around the current posture was not learned yet.
  During the update of MRCM, RMSE decreased slowly.
  As stated in \secref{subsec:mrcm}, the model of MRCM is simplified and trained, so the obtaining of correct weight takes a long time.
  We show the comparison of IJMM between before and after online learning in \figref{figure:learning-model}.
  Also, we show the comparison of MRCM between before and after online learning in \figref{figure:learning-emodel}.
  We can see the MRCM was modified from the original geometric model by online learning.
  Also, at the deltoid (front) and triceps brachii, we can see the nonlinear spring feature of the actual robot muscles was learned.
}%
{%
  本節ではIJMMとMRCMの逐次的再学習の結果を示す.
  先行研究\cite{ral2018:kawaharazuka:vision-learning}でAntagonism Updaterの有効性は示されているため, ここではIJMMとMRCMのVision Updaterの有効性を示す.

  実験の様子は図\figref{figure:learning-experiment}の上段に示されている.
  はじめに腱悟郎に任意の関節角度指令を送り, その際に自身の体を目で見ることでVision Updaterを用いて自己身体像を更新する.
  この際は$\bm{l}_{target} != constant$なため, IJMMが更新される.
  次に, 筋長指令を固定した状態($\bm{l}_{target} = constant$)で外力を加えることで, MRCMを更新していく.
  なお, 全ての動作中にAntagonism Updaterも動作している.

  この際の首・肩甲骨・肩・前腕のそれぞれの関節群における, 視覚から得た実機の関節角度$\bm{\theta}_{actual}$と現在の自己身体像から計算された関節角度推定値$\bm{\theta}_{est}$の差分のRMSE(Root Mean Squared Error)の推移を図\figref{figure:learning-experiment}の下段に示す(以降RMSEと言うとこれを指す).
  IJMMの更新期間では, はじめは16deg程度だったRMSEが, 最終的に0$\sim$2deg程度まで下がっていることがわかる.
  また, 20secでもう一度RMSEが上昇しているのが分かるが, これは0$\sim$5secで学習された姿勢から大きく動作したため, まだその姿勢周辺の学習が進んでいなかったからである.
  MRCMの更新期間では, ゆっくりではあるが, 少しずつRMSEが下がっているのがわかる.
  MRCMは\figref{subsec:mrcm-model}節で述べたようにモデルを単純化して初期学習を行っているため, 正しい重みを構築するのに時間がかかると考えられる.
}%

\begin{figure}[t]
  \centering
  \includegraphics[width=1.0\columnwidth]{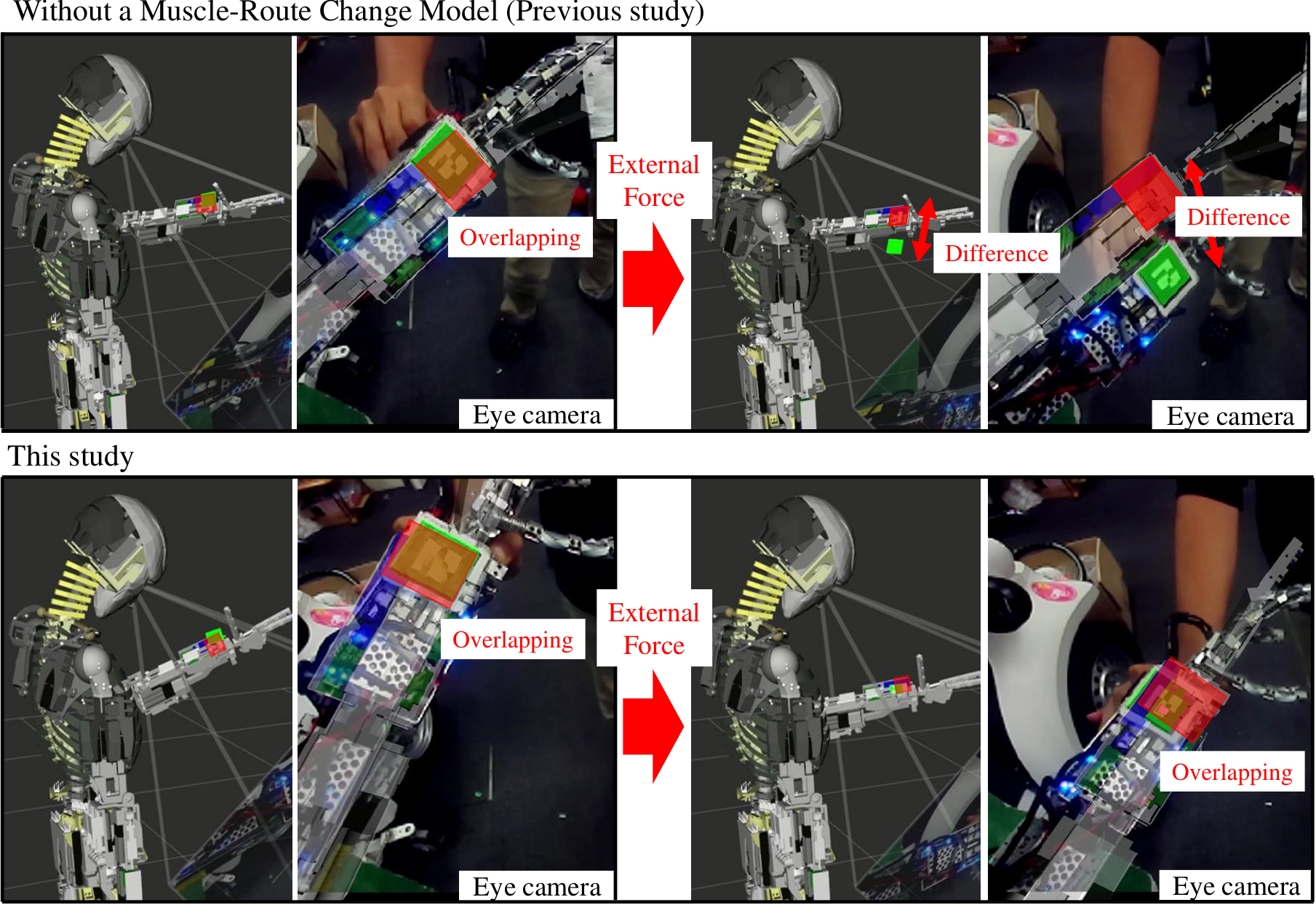}
  \caption{Experiment of joint angle estimator: upper figure shows the joint angle estimation result in the previous study \cite{ral2018:kawaharazuka:vision-learning} and lower shows that of this study.}
  \label{figure:estimator-experiment}
\end{figure}

\subsection{Joint Angle Estimation}
\begin{figure}[t]
  \centering
  \includegraphics[width=1.0\columnwidth]{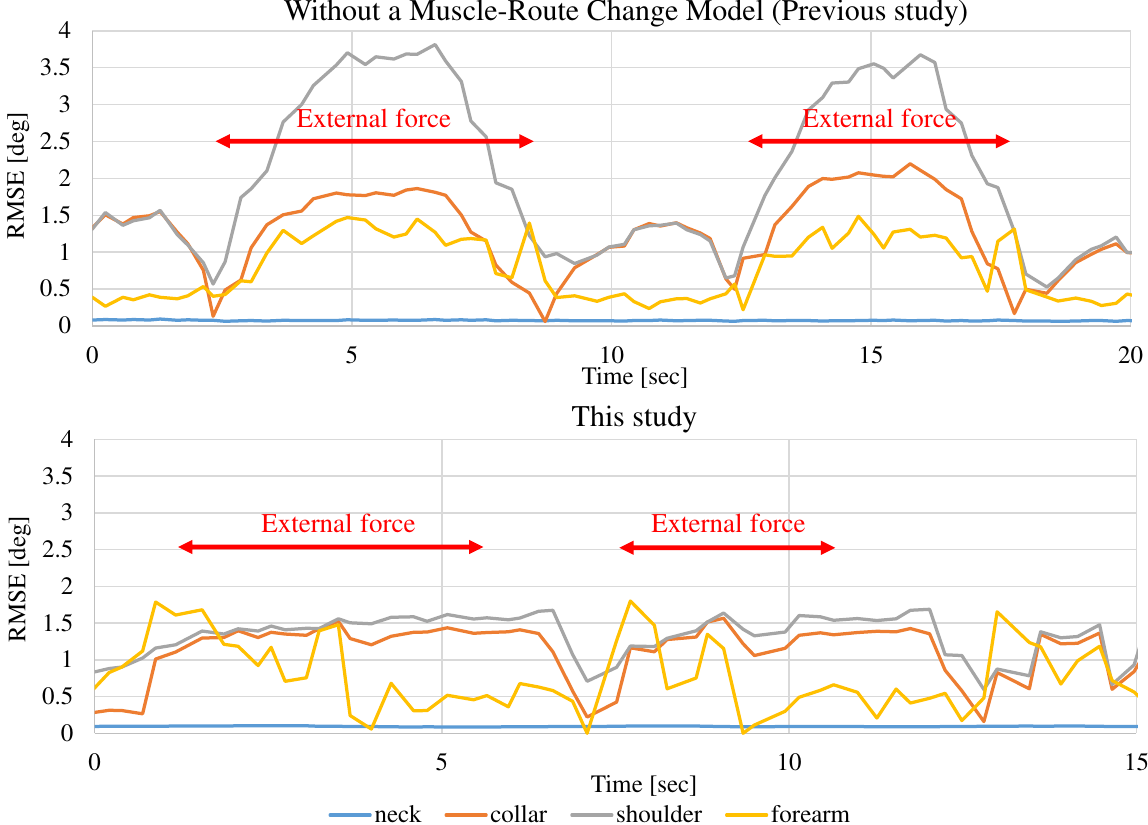}
  \caption{Transition of RMSE during joint angle estimation: previous study \cite{humanoids2015:ookubo:muscle-learning} and this study.}
  \label{figure:estimator-experiment-graph}
\end{figure}

\switchlanguage%
{%
  We compared the joint angle estimation using self-body image acquisition without MRCM conducted in a previous study \cite{ral2018:kawaharazuka:vision-learning} and the joint angle estimation using the proposed self-body image in this study.
  Under the condition that Kengoro keeps the posture, we applied external force and compared the change of the estimated joint angles.
  In this experiment, self-body image is learned previously and is not learned during this experiment.

  The result of the experiment is shown in \figref{figure:estimator-experiment}.
  In the previous study (upper figure of \figref{figure:estimator-experiment}), when the external force is applied, the estimated joint angles did not follow the actual joint angles, and there was a large error between self-body image and the actual robot.
  In comparison, in this study (lower figure of \figref{figure:estimator-experiment}), even when the external force is applied, the estimated joint angles followed the actual joint angles, and better self-body image was acquired.
  We show the comparison of the transition of RMSE between the previous study and this study in \figref{figure:estimator-experiment-graph}.
  In the previous study, when the external force was applied, the RMSE of the shoulder rose up to about 4 deg, but in this study, the RMSE was within 2 deg.
}%
{%
  本節では, 先行研究\cite{ral2018:kawaharazuka:vision-learning}で行ったMRCMのない自己身体像を用いた関節角度推定と, 本研究の自己身体像を用いた関節角度推定の比較を行う.
  それぞれ腱悟郎の姿勢を固定した状態で, 外力を加えた際の関節角度推定の変化を比較した.
  なお, 以降の実験ではそれぞれの自己身体像は先に学習済みであり, 実験中は逐次的再学習は行っていない.

  実験結果の様子は図\figref{figure:estimator-experiment}であり, 先行研究(上段)では外力が働いた際に関節角度推定が追従せず, 実機と大きな差異があることがわかる.
  それに対して, 本研究(下段)では, 外力が働いた場合でも関節角度推定が追従し, より正しい自己身体像が得られていることがわかる.
  この際の先行研究と本研究でのRMSEの推移を比較したグラフが図\figref{figure:estimator-experiment-graph}である.
  先行研究では外力を加えると肩のRMSEが4deg程度まで上昇しているが, 本研究では2deg以内で収まっていることが分かる.
}%

\subsection{Heavy Object Grasping}
\begin{figure}[t]
  \centering
  \includegraphics[width=1.0\columnwidth]{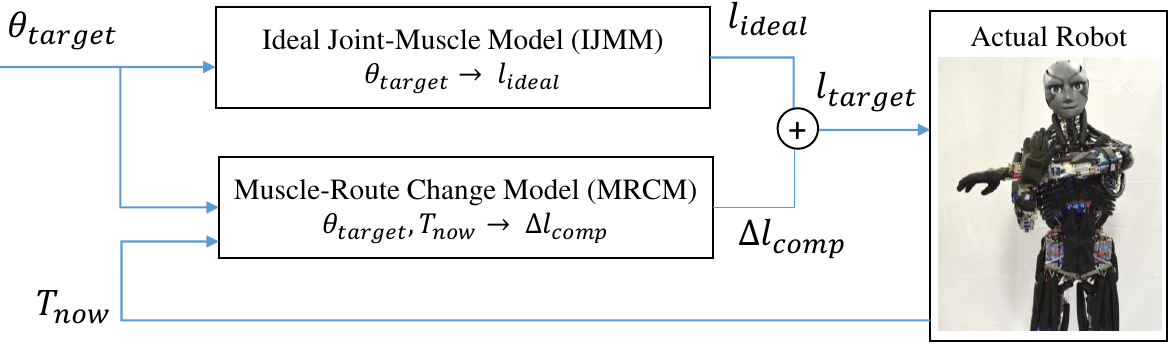}
  \caption{A control using self-body image proposed in this study. Feedback control of joint angles using IJMM, MRCM, and the actual robot muscle tensions.}
  \label{figure:dumbbell-control}
\end{figure}

\begin{figure}[t]
  \centering
  \includegraphics[width=1.0\columnwidth]{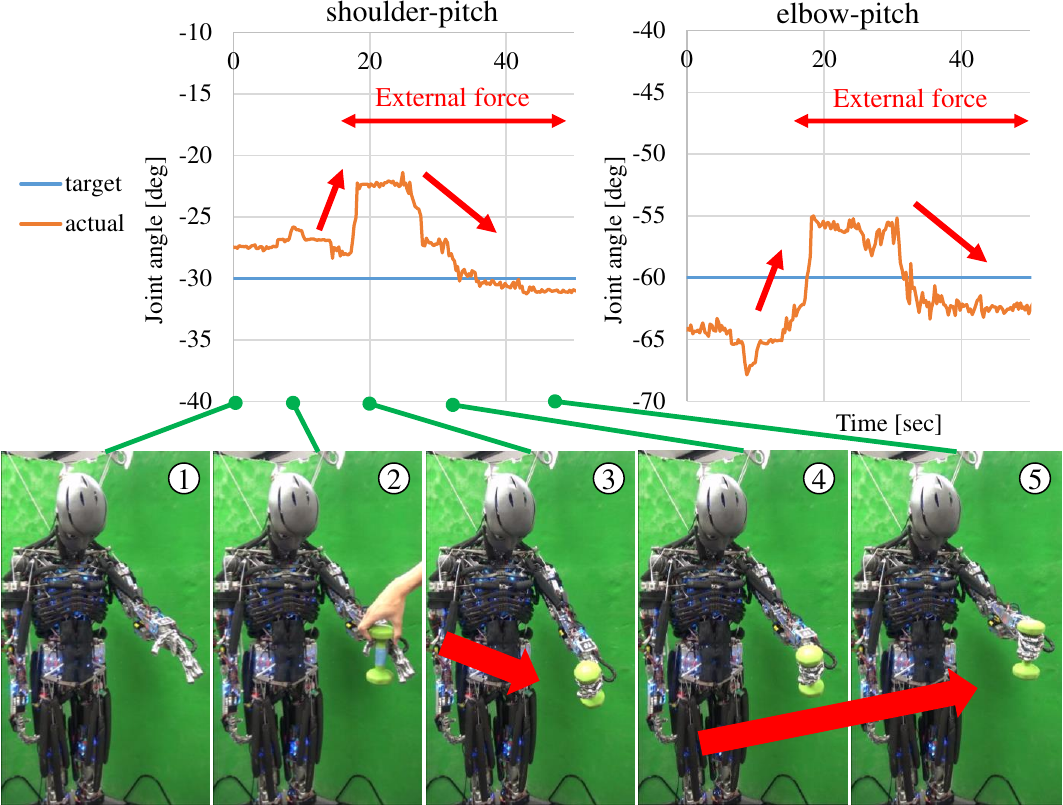}
  \caption{Experiment of dumbbell grasping.}
  \label{figure:dumbbell-experiment}
\end{figure}

\switchlanguage%
{%
  Finally, we show the control using self-body image which is learned in this study (\figref{figure:dumbbell-control}).
  We conducted an experiment in which the robot grasps a heavy object (in this experiment, we used 1.4 kg dumbbell) and continues to keep the target posture.
  We show the result of the experiment and the graph which compares the target joint angles $\bm{\theta}_{target}$ and the actual joint angles $\bm{\theta}_{actual}$ regarding the elbow and shoulder-pitch joints in \figref{figure:dumbbell-experiment}.

  First, the robot moves to the posture in which the shoulder pitch is -30 deg and the elbow is -60 deg, and grasps the dumbbell.
  The arm of the robot fell down because of the muscle route changes caused by softness of body tissue.
  However, by embedding a closed loop control to the change of muscle tension, which calculates compensation value of muscle length from current muscle tension and sends it to the robot, the posture of the robot went back to the original posture.
  In this experiment, the frequency of feedback control is low and the period of the compensation is long for better visual understandability.
}%
{%
  %最後に, 本研究で得られた自己身体像を用いた制御を一つ示す.
  %重量物体(本実験では約1.4kgのダンベル)をターゲットとなる姿勢で保持し続ける実験を行った.
  %その際の実験の様子, 肩pitchと肘pitchのターゲットとなる関節角度と実機の関節角度の比較グラフを図\figref{figure:dumbbell-experiment}に示す.

  %はじめに腱悟郎の肩pitchを-30deg, 肘pitchを-60degとし, その状態でダンベルを握らせる.
  %すると身体組織の柔軟性による筋経路変化によって腕は下方へ下がるが, この際に現在の張力から筋経路変化補償分の筋長を算出しそれを実機に送るという筋張力変化に対する制御ループを組むことで, 最終的に元の位置まで姿勢が修正されているのがわかる.
}%

%%%%%%%%%%%%%%%%%%%%%%%%%%%%%%%%%%%%%%%%%%%%%%%%%%%%%%%%%%%%%%%%%%%%%%%%%%%%%%%%
\section{CONCLUSION} \label{sec:6}
\switchlanguage%
{%
  In this study, we proposed an online learning method of self-body image considering muscle route changes caused by softness of body-tissue.
  In tendon-driven musculoskeletal humanoids, because of their body complexity, they are difficult to modelize, and there are several problems in controllability such as challenges in moving the actual robot as intended in a simulation environment, emergence of large internal muscle tension, and existence of movements which do not appear as changes in muscle lengths.
  To solve these problems, we constructed self-body image using a neural network and updated it online using the sensor information of the actual robot.
  To consider the muscle route changes which do not appear as changes in muscle lengths, we constructed self-body image that calculates target muscle lengths from the target joint angles and muscle tensions.
  Also, to learn self-body image correctly, we divided self-body image into two models: ideal joint-muscle model and muscle-route change model.
  First, we initialized the two models using a man-made geometric model.
  Second, we updated them online using two updaters: Antagonism Updater and Vision Updater to modify the model error between the actual robot and its geometric model.
  Finally, we showed effectiveness of this proposed system by the experiments of the joint angle estimation and grasping of a heavy object.

  In future works, we would like to make the online learning of self-body image more stable.
  In order to make this method practical, we will consider how efficiently self-body image can be obtained using little actual robot sensor information.
  Also, we dealt with only the static state in this study, so next, we would like to deal with self-body image in a dynamic state.
}%
{%
  本研究では, 筋骨格ヒューマノイドにおける、身体組織の柔軟性による筋経路変化を考慮した自己身体像の逐次的再学習法を提案した.
  筋骨格ヒューマノイドはその構造複雑性ゆえに、モデル化が難しく、シミュレーション上で意図した動きが実現できない・大きな内力が発生する・筋長変化には現れない関節の動きが存在する等の制御的問題がある。
  その中でも、筋長センサの変化として現れない筋経路変化を考慮するためには筋張力を考慮する必要があり, ターゲットとなる姿勢と筋張力から送るべき筋長を算出するモデルをニューラルネットワークを用いて構成し、それをオンラインで更新する手法を考案した.
  その際学習を正しく進ませるために, 自己身体像を理想的な関節-筋空間モデルと筋経路変化に対する補償モデルに分けることで問題を切り分けた.
  まず二つのモデルを人間の作った幾何モデルから初期学習させ、ある程度のモデルを構築する。
  さらに幾何モデルと実機の間の誤差を修正するために、Antagonism UpdaterとVision Updaterを考案し二つのモデルの同時更新則を考案した。
  最後に、本手法の有効性を関節角度推定・重量物体保持において示した.

  今後の展望としてはまず, 自己身体像の学習の安定化が挙げられる.
  より実用的なものとするため, 過学習せず, 少ないデータから如何に効率的に自己身体像を獲得するかが鍵となる.
  また, 本研究では静的な状態のみ扱っているため, 今後はより動的な自己身体像の獲得も扱っていきたい.
}%

{
  %\footnotesize
  %\small
  %\bibliographystyle{junsrt}
  \bibliographystyle{IEEEtran}
  \bibliography{main}
}

\end{document}